\def\year{2021}\relax
\documentclass[letterpaper]{article} 
\usepackage{aaai21}  
\usepackage{times}  
\usepackage{helvet} 
\usepackage{courier}  
\usepackage[hyphens]{url}  
\usepackage{graphicx} 
\usepackage{natbib}  
\usepackage{caption} 
\usepackage{enumitem}
\usepackage{multirow}
\usepackage{booktabs}
\usepackage{subcaption}
\usepackage{xr}
\usepackage{bbold}
\usepackage{booktabs}
\usepackage{amsmath}
\usepackage{xcolor,pifont}
\usepackage{MnSymbol,wasysym}
\usepackage{fontawesome}
\usepackage{graphicx}
\usepackage{anyfontsize}
\usepackage{tikz}
\usepackage{todonotes}
\usepackage{array}
\newcommand\newtag[2]{#1\def\@currentlabel{#1}\label{#2}}

\title{A Paragraph-level Multi-task Learning Model for Scientific Fact-Verification}
\author {
    Xiangci Li,\textsuperscript{\rm 1} \thanks{\ \  Work performed at Information Sciences Institute, Viterbi School of Engineering, University of Southern California}
    Gully Burns, \textsuperscript{\rm 2}
    Nanyun Peng \textsuperscript{\rm 3} \\
}
\affiliations {
    \textsuperscript{\rm 1} University of Texas at Dallas \\
    \textsuperscript{\rm 2} Chan Zuckerburg Initiative \\
    \textsuperscript{\rm 3} University of California Los Angeles \\
    lixiangci8@gmail.com, gully.burns@chanzuckerburg.com, violetpeng@cs.ucla.edu
}

\begin{document}

\maketitle

\begin{abstract}
Even for domain experts, it is a non-trivial task to verify a scientific claim by providing supporting or refuting evidence rationales. The situation worsens as misinformation is proliferated on social media or news websites, manually or programmatically, at every moment. As a result, an automatic fact-verification tool becomes crucial for combating the spread of misinformation. 
In this work, we propose a novel, paragraph-level, multi-task learning model for the \textsc{SciFact} task by directly computing a sequence of contextualized sentence embeddings from a BERT model and jointly training the model on rationale selection and stance prediction.
\end{abstract}

\section{Introduction}
Many seemingly convincing rumors such as ``Most humans only use 10 percent of their brain'' are widely spread, but ordinary people are not able to rigorously verify them by searching for scientific literature. In fact, it is not a trivial task to verify a scientific claim by providing supporting or refuting evidence rationales, even for domain experts. 
The situation worsens as misinformation is proliferated 
on social media or news websites, manually or programmatically, at every moment. As a result, an automatic fact-verification tool becomes more and more crucial for combating 
the spread of misinformation.

The existing fact-verification tasks usually consist of three sub-tasks: document retrieval, rationale sentence extraction, and fact-verification. However, due to the nature of scientific literature that requires domain knowledge, it is challenging to collect a large scale scientific fact-verification dataset, and further, to perform fact-verification under a low-resource setting with limited training data. \citet{Wadden2020FactOF} collected a scientific claim-verification dataset, \textsc{SciFact}, and proposed a scientific claim-verification task: given a scientific claim, find evidence sentences that support or refute 
the claim 
in a corpus of scientific paper abstracts. \citet{Wadden2020FactOF} also proposed a simple, pipeline-based, sentence-level model, \textsc{VeriSci}, as a baseline solution based on \citet{deyoung2019eraser}.

\textsc{VeriSci} is a pipeline model that runs modules for abstract retrieval, rationale sentence selection, and stance prediction sequentially, and thus the error generated from 
an upstream module may propagate to the downstream modules. To overcome this drawback, we hypothesize that a module jointly optimized on multiple sub-tasks may mitigate the error-propagation problem to improve the overall performance. 
In addition, we observe that a complete set of rationale sentences usually contains multiple inter-related sentences from the same paragraph. Therefore, we propose a novel, paragraph-level, multi-task learning model for the \textsc{SciFact} task.

In this work, we employ \textit{compact paragraph encoding}, a novel strategy of computing sentence representations using BERT-family models. We directly feed an entire paragraph as a single sequence to BERT, so that the encoded sentence representations are already contextualized on the neighbor sentences by taking advantage of the attention mechanisms in BERT. In addition, we jointly train the modules for rationale selection and stance prediction as multi-task learning \cite{caruana1997multitask} by leveraging the confidence score of rationale selection as the attention weight of the stance prediction module. Furthermore, we compare two methods of transfer learning that mitigate the low-resource issue: pre-training and domain adaptation \cite{peng2016multi}. Our experiments show that: 

\begin{itemize}[leftmargin=*]
    \item The \textit{compact paragraph encoding} method is beneficial over separately computing sentence embeddings.
    \item With negative sampling, the joint training of rationale selection and stance prediction is beneficial over the pipeline solution. 

\end{itemize}

\section{\textsc{SciFact} Task Formulation}
Given a scientific claim $c$ and a corpus of scientific paper abstracts $A$, the \textsc{SciFact} \cite{Wadden2020FactOF} task retrieves all abstracts $\hat{E}(c)$ that either \textsc{Supports} or \textsc{Refutes} $c$. Specifically, the \emph{stance prediction} (a.k.a. \emph{label prediction}) task classifies each abstract $a \in A$ into $y(c,a) \in \{\text{\textsc{Support}}, \text{\textsc{Refutes}}, \text{\textsc{NoInfo}}\}$ with respect to each claim $c$;
the \emph{rationale selection} (a.k.a. \emph{sentence selection}) task retrieves all rationale sentences $\hat{S}(c,a)=\{\hat{s_1}(c,a),...,\hat{s_l}(c,a)\}$ of each $a$ that \textsc{Supports} or \textsc{Refutes} $c$. The performance of both tasks are evaluated with $F1$ measure 
at both abstract-level and sentence-level, as defined by \citet{Wadden2020FactOF}, where $\{\text{\textsc{Supports}}, \text{\textsc{Refutes}}\}$ are considered as the positive labels and \textsc{NoInfo} is the negative label for \emph{stance prediction}.

\section{Approach}
We formulate the \textsc{SciFact} task \cite{Wadden2020FactOF} as a sentence-level sequence-tagging problem. We first apply an \emph{abstract retrieval} module to filter out negative candidate abstracts that do not contain sufficient information with respect to each given claim. Then we propose a novel model for joint \emph{rationale selection} and \emph{stance prediction} using multi-task learning \cite{caruana1997multitask}.

\subsection{Abstract Retrieval}
In contrast to the TF-IDF similarity used by \citet{Wadden2020FactOF},  we leverage BioSentVec \cite{chen2019biosentvec} embedding, which is the biomedical version of Sent2Vec \cite{pgj2017unsup}, for a fast and scalable sentence-level similarity computation. We first compute the BioSentVec \cite{chen2019biosentvec} embedding of each abstract in the 
corpus by treating the concatenation of each title and abstract as a single sentence. Then for each given claim, we compute the cosine similarities of the claim embedding against the pre-computed abstract embeddings, and choose the top $k_{retrieval}$ similar abstracts as the candidate abstracts for the next module. 

\subsection{Joint Rationale Selection and Stance Prediction Model}
\subsubsection{Compact Paragraph Encoding}
\label{paragraph_encoding}
A major usage of BERT-family models \cite{devlin2018bert, liu2019roberta} for sentence-level sequence tagging computes each sentence embedding in a paragraph with batches. Since each batch is independent, such method leaves the contextualization of the sentences to the subsequent modules. Instead, we propose a novel method of encoding paragraphs by directly feeding the concatenation of the claim $c$ and the whole paragraph $P$ to a BERT model $BERT$ as a single sequence $Seq$. By separating each sentence $s$ using the BERT model's $[SEP]$ token, we fully leverage the multi-head attention \cite{vaswani2017attention} within the BERT model to compute the contextualized word representations $h_{Seq}$ with respect to the claim sentence and the whole paragraph. 

\begin{equation}
\small
\begin{aligned}
c &= [cw_1, cw_2, \ldots, cw_n] \\
s_i &= [w_1, w_2, \ldots, w_m] \\
P &= [s_1, s_2, \ldots, s_l] \\
Seq &= [c[SEP]s_1[SEP]s_2[SEP]\ldots[SEP]s_l] \\
h_{Seq} &= BERT(Seq) \in R^{len(Seq) \times d_{BERT}}\\
h_{Seq} &= [h_{CLS}, {h_{cw}}_1, \ldots, {h_{cw}}_n, \\ 
& h_{SEP}, {h_w}_1, \ldots, {h_w}_m, h_{SEP},\ldots] \\
\end{aligned}
\end{equation}

\subsubsection{Sentence Representations via Word-level Attention}
\label{word_attention}
Next, we apply a weighted sum to the contextualized word representations of each sentence $h_{sent}$ to compute the sentence representations $h_{s_{i}}$. The 
weights are obtained by applying a self-attention 
$SelfAttn_{word}$ with a two-layer multi-layer perceptron on the word representations in the scope of each sentence, as separated by the $[SEP]$ tokens.

\begin{equation}
\small
\begin{aligned}
h_{s_{i}} &= SelfAttn_{word}([h_{SEP}, h_{w1}, ..., h_{wm}]) \in R^{d_{BERT}}
\end{aligned}
\end{equation}

\subsubsection{Dynamic Rationale Representations}
\label{rationale_selection}
We use a two-layer multi-layer perceptron $MLP_{rationale}$ to compute the rationale score and use the $softmax$ function to compute the probability of each candidate sentence being a rationale sentence $p^{r}$ or not $p^{not\_r}$ with respect to the claim sentence $c$. Then we only feed rationale sentences $r$ into the next stance prediction module.

\begin{equation}
\small
\begin{aligned}
p^{not\_r}_i, p^{r}_i &= softmax(MLP_{rationale}({h_s}_i)) \in (0,1) \\
{h_r}_i &\leftarrow {h_s}_i \text{ if } p^{not\_r}_i < p^{r}_i
\end{aligned}
\end{equation}

\subsubsection{Stance Prediction}
\label{stance_prediction}
We use two variants for \emph{stance prediction}: a simple sentence-level attention
and the Kernel Graph Attention Network (KGAT) \cite{liu2020fine}.
\begin{itemize}[leftmargin=*]
    \item \textbf{Simple Attention.}
    We apply another weighted summation on the predicted rationale sentence representations $h_{r_i}$ to compute the whole paragraph's rationale representation, where the attention weights are obtained by applying another self-attention 
    $SelfAttn_{sentence}$ on the rationale sentence representations $h_r$. Finally, we apply another two-layer multi-layer perceptron $MLP_{stance}$ and the $softmax$ function to compute the probability of the paragraph 
    serving the role of \{\textsc{Supports}, \textsc{Refutes}, \textsc{NoInfo}\} with respect to the claim $c$.

\begin{equation}
\small
\begin{aligned}
h_{r} &= SelfAttn_{sentence}([h_{r1}, h_{r2},..., h_{rl}]) \in R^{d_{BERT}} \\
p^{stance} &= softmax(MLP_{stance}(h_{r})) \in (0,1)^3
\end{aligned}
\end{equation}

    \item \textbf{Kernel Graph Attention Network.}
    \citet{liu2020fine} proposed KGAT as a \emph{stance prediction} module for their pipeline solution on the FEVER \cite{thorne2018fever} task. In addition to the Graph Attention Network \cite{velivckovic2017graph}, which applies attention mechanisms on each word pair and 
    sentence pair in the input paragraph, KGAT applies a kernel pooling mechanism \cite{xiong2017end} 
    to extract better features for \emph{stance prediction}. We integrate KGAT \cite{liu2020fine} into our multi-task learning model for \emph{stance prediction} on \textsc{SciFact} \cite{Wadden2020FactOF}. The KGAT module $KGAT$ takes the word representation of the claim $h_c$ and the predicted rationale sentence representations $h_{R}$ as inputs, and outputs the probability of the paragraph 
    serving the role of \{\textsc{Supports}, \textsc{Refutes}, \textsc{NoInfo}\} with respect to the claim $c$.

\begin{equation}
\small
\begin{aligned}
h_{c} &= [h_{CLS}, {h_cw}_1, \ldots, {h_cw}_n] \\
{h_R}_i &=  [h_{SEP}, {h_rw}_1, \ldots, {h_rw}_m] \text{ where } p^{not\_r}_i < p^{r}_i \\
h_{R} &= [{h_R}_1, {h_R}_2, \ldots, {h_R}_l] \\
p^{stance} &= KGAT(h_{c}, h_{R}) \in (0,1)^3
\end{aligned}
\end{equation}

\end{itemize}

\subsection{Model Training}

\subsubsection{Multi-task Learning}
We train our model on \emph{rationale selection} and \emph{stance prediction} using multi-task learning approach \cite{caruana1997multitask}. We use cross-entropy loss as 
the training objective for both tasks. We introduce a coefficient $\gamma$ to adjust the proportion of two loss values $L_{rationale}$ and $L_{stance}$ in the joint loss $L$.

\begin{equation}
\small
\begin{aligned}
L &= \gamma L_{rationale} + L_{stance}
\end{aligned}
\end{equation}

\subsubsection{Scheduled Sampling}
Because the \emph{stance prediction} module takes the predicted rationale sentences as the input, 
errors 
in rationale selection may propagate to the \emph{stance prediction} module, especially during the early stage of training. To mitigate this issue, we apply scheduled sampling \cite{bengio2015scheduled}, which starts 
by feeding the \emph{ground truth} rationale sentences to the stance prediction module, and gradually increasing the proportion of the predicted rationale sentences, until eventually all input sentences are the predicted rationale sentences. We use a $sin$ function to compute the probability of sampling predicted rationale sentences $p_{sample}$ as a function of the progress of the training:

\begin{equation}
\small
\begin{aligned}
progress &= \frac{current\_epoch - 1}{total\_epoch - 1}  \\
p_{sample} &= sin(\frac{\pi}{2} \times progress)
\end{aligned}
\end{equation}

\subsubsection{Negative Sampling and Down-sampling}
Although the \emph{abstract retrieval} module filters out the majority of the negative candidate abstracts, the false-positive rate is still inevitably high, in order to ensure the retrieval of most of the positive abstracts. As a result, the input to the joint prediction model is highly biased towards negative samples. Therefore, in addition to the positive samples from the \textsc{SciFact} dataset \cite{Wadden2020FactOF}, we perform negative sampling \cite{mikolov2013distributed} to sample the top $k_{train}$ similar negative abstracts 
using our \emph{abstract retrieval} module as an augmented dataset for training and validation to increase the downstream model's tolerance to false positive abstracts. Furthermore, in order to increase the 
diversity of the dataset, we augment the dataset by down-sampling sentences within each paragraph.

\subsubsection{FEVER Pre-training}
As \citet{Wadden2020FactOF} proposed, due to the similar task structure of FEVER \cite{thorne2018fever} and \textsc{SciFact} \cite{Wadden2020FactOF}, we first pre-train our model on the FEVER dataset, then fine-tune on the \textsc{SciFact} dataset by partially re-initializing the \emph{rationale selection} and \emph{stance prediction} attention modules.

\subsubsection{Domain Adaptation}
Instead of pre-training, we also explore 
domain adaptation \cite{peng2016multi} from FEVER \cite{thorne2018fever} to \textsc{SciFact} \cite{Wadden2020FactOF}. We use shared representations for the \emph{compact paragraph encoding} and word-level attention, while using domain-specific representations for the \emph{rationale selection} and \emph{stance prediction} modules.

\subsection{Implementation Details}
\paragraph{BERT Encoding.} We follow \citet{Wadden2020FactOF} 
in using Roberta-large \cite{liu2019roberta} as our BERT-family model.

\paragraph{Dummy Rationale Sentence.} We dynamically feed only the predicted rationale sentence representations to the \emph{stance prediction} module. To address the special case when an abstract contains no rationale sentences, we append a fixed dummy sentence (e.g.``@'') whose rationale label is always $0$ at the beginning of each of the paragraph. When the \emph{stance prediction} module has no actual rationale sentence to take as input, we feed it with the representation of the dummy sentence and expect the module to predict \textsc{NoInfo}.

\paragraph{Post Processing.} To 
prevent inconsistency between the outputs of rationale selection and stance prediction, we enforce the predicted stance to be \textsc{NoInfo} if no rationale sentence is proposed.

\paragraph{Hyper-parameters.}
Table \ref{tab:hyper} lists the hyper-parameters used for training the Joint-Paragraph model in Table \ref{tab:open_dev} \footnote{\url{https://github.com/jacklxc/ParagraphJointModel}}, where $k_{FEVER}$ refers to the number of negative samples retrieved from FEVER \cite{thorne2018fever} for model pre-training. 

\begin{table}[t]
\begin{center}
\begin{tabular}{c|c|c}
 \hline
Parameter & Explored &  Used \\  \hline
$k_{retrieval}$ &$3 \sim 100$ &$30$ \\ \hline
$k_{FEVER}$ &$1 \sim 15$ &$5$ \\ \hline
$k_{train}$ &$0 \sim 50$ &$12$ \\ \hline
$\gamma$ &$0.1 \sim 10$ &$6$ \\ \hline
drop out &$0 \sim 0.6$ &$0$ \\ \hline
learning rate &$1 \times 10^{-6} \sim 1 \times 10^{-4}$ &$5 \times 10^{-6}$ \\ \hline
BERT learning rate &$1 \times 10^{-6} \sim 1 \times 10^{-4}$ &$1 \times 10^{-5}$ \\ \hline
batch size &$ 1, 2 $ &$1$ \\ \hline
\end{tabular}
 \caption{Hyper-parameters explored and used.}
   \label{tab:hyper}
 \end{center}
\end{table}

\begin{table}[t]
\begin{center}
\begin{tabular}{c|ccc|ccc}
 \hline
& \multicolumn{3}{c}{TF-IDF} &  \multicolumn{3}{c}{BioSentVec} \\ 
$k_{retrieval}$ & P & R & $F_1$ & P & R & $F_1$ \\ \hline
$3$ &$16.2$ &$69.9$ &$26.3$ &$15.6$ &$67.0$  &$25.3$ \\ \hline
$10$ &$5.83$ &$83.6$ &$10.9$ &$5.86$ &$84.2$ &$11.0$ \\ \hline
$100$ &$0.67$ &$96.7$ &$1.33$ &$0.68$ &$98.1$ &$1.35$ \\ \hline
$150$ &$0.45$ &$96.7$ &$0.90$ &$0.46$ &$98.1$ &$0.92$ \\ 

 \hline
 \hline

\end{tabular}
 \caption{Abstract retrieval performance on dev set in \%.}
   \label{tab:abstract_retrieval}
 \end{center}
\end{table}

\begin{table*}[t]
\begin{center}
\begin{tabular}{m{11em}|m{1.5em}m{1.5em}m{1.8em}|m{1.5em}m{1.5em}m{1.8em} | m{1.5em}m{1.5em}m{1.8em}|m{1.5em}m{1.5em}m{1.8em}}
 \hline
 & \multicolumn{6}{c}{Sentence-level} &  \multicolumn{6}{c}{Abstract-level}\\ 
& \multicolumn{3}{c}{Selection-Only} &  \multicolumn{3}{c}{Selection+Label} & \multicolumn{3}{c}{Label-Only} &  \multicolumn{3}{c}{Label+Rationale}\\ 
Models & P & R & $F_1$ & P & R & $F_1$ & P & R & $F_1$ & P & R & $F_1$\\ \hline
\textsc{VeriSci} &$77.7$ &$56.3$ &$65.3$ &$69.8$ &$50.5$ &$58.6$ &$89.6$ &$66.0$ &$76.0$ &$84.4$ &$62.2$ &$71.6$ \\ \hline
 \hline
Paragraph-Pipeline &$74.9$ &$67.7$ &$71.2$ &$65.6$ &$59.3$ &$62.3$ &$81.6$ &$72.2$ &$76.6$ &$76.8$ &$67.9$ & $72.1$\\ \hline
Paragraph-Joint &$75.9$ &$62.8$ &$68.8$ &$67.7$ &$56.0$ &$61.3$ &$83.3$ &$76.6$ &$\textbf{79.8}$ &$76.6$ &$70.3$ &$\textbf{73.3}$ \\
Paragraph-Joint KGAT &$75.5$ &$68.3$ &$\textbf{71.7}$ &$66.2$ &$59.8$ &$\textbf{62.8}$ &$81.8$ &$75.1$ &$78.3$ &$76.5$ &$70.3$ &$73.3$ \\ 
 \hline
 \hline
\textsc{VerT5erini}* &$83.5$ &$72.1$ &$77.4$ &$78.2$ &$67.5$ &$72.4$ &$92.7$ &$79.0$ &$85.3$ &$88.8$ &$75.6$ &$81.7$ \\ \hline
\end{tabular}
 \caption{Model performance on dev set oracle abstracts in \%. The model with * is only for reference.}
   \label{tab:oracle_dev}
 \end{center}
\end{table*}

\begin{table*}[t]
\begin{center}
\begin{tabular}{m{12em}|m{1.5em}m{1.5em}m{1.8em}|m{1.5em}m{1.5em}m{1.8em} | m{1.5em}m{1.5em}m{1.8em}|m{1.5em}m{1.5em}m{1.8em}}
 \hline
 & \multicolumn{6}{c}{Sentence-level} &  \multicolumn{6}{c}{Abstract-level}\\ 
& \multicolumn{3}{c}{Selection-Only} &  \multicolumn{3}{c}{Selection+Label} & \multicolumn{3}{c}{Label-Only} &  \multicolumn{3}{c}{Label+Rationale}\\ 
Models & P & R & $F_1$ & P & R & $F_1$ & P & R & $F_1$ & P & R & $F_1$\\ \hline
\textsc{VeriSci} &$54.3$ &$43.4$ &$48.3$ &$48.5$ &$38.8$ &$43.1$ &$56.4$ &$48.3$ &$52.1$ &$54.2$ &$46.4$ &$50.0$ \\ \hline
 \hline
Paragraph-Pipeline &$71.2$ &$51.4$ &$59.7$ &$62.1$ &$44.8$ &$52.1$ &$77.6$ &$54.5$ &$64.0$ &$72.8$ &$51.2$ &$60.1$\\ \hline
Paragraph-Joint  &$74.2$ &$57.4$ &$\textbf{64.7}$ &$63.3$ &$48.9$ &$\textbf{55.2}$ &$71.4$ &$59.8$ &$65.1$ &$65.7$ &$55.0$ &$59.9$ \\
Paragraph-Joint SciFact-only &$69.3$ &$50.0$ &$58.1$ &$59.8$ &$43.2$ &$50.2$ &$69.9$ &$52.1$ &$59.7$ &$64.7$ &$48.3$ &$55.3$ \\
Paragraph-Joint TF-IDF &$72.5$ &$55.7$ &$63.1$ &$62.2$ &$47.8$ &$54.1$ &$70.6$ &$59.8$ &$64.7$ &$65.5$ &$55.5$ &$60.1$ \\
Paragraph-Joint DA &$69.4$ &$56.6$ &$62.3$ &$60.4$ &$49.2$ &$54.2$ &$67.4$ &$57.4$ &$62.0$ &$63.5$ &$54.1$ &$58.4$ \\ \hline
Paragraph-Joint KGAT &$68.8$ &$56.6$ &$62.1$ &$60.1$ &$49.5$ &$54.3$ &$68.2$ &$62.7$ &$\textbf{65.3}$ &$61.5$ &$56.5$ &$58.9$ \\ 
Paragraph-Joint KGAT DA &$70.2$ &$55.5$ &$62.0$ &$61.9$ &$48.9$ &$54.7$ &$70.9$ &$59.3$ &$64.6$ &$66.3$ &$55.5$ &$\textbf{60.4}$ \\ 
 \hline
 \hline
\textsc{VerT5erini} (BM25)* &$67.7$ &$53.8$ &$60.0$ &$63.9$ &$50.8$ &$56.6$ &$70.9$ &$61.7$ &$66.0$ &$67.0$ &$58.4$ &$62.4$ \\
\textsc{VerT5erini} (T5)* &$64.8$ &$57.4$ &$60.9$ &$60.8$ &$53.8$ &$57.1$ &$65.1$ &$65.1$ &$65.1$ &$61.7$ &$61.7$ &$61.7$ \\ \hline

\end{tabular}
 \caption{Model performance on dev set with abstract-retrieval in \%. The models with * are only for reference.}
   \label{tab:open_dev}
 \end{center}
\end{table*}

\section{Experiments}

\subsection{\textsc{SciFact} Dataset}
\textsc{SciFact} \cite{Wadden2020FactOF} is a small dataset, whose corpus contains 5183 abstracts. There are 1409 claims, including 809 in the training set, 300 in the development set and 300 in the test set. 

\subsection{Abstract Retrieval Performance}
Table \ref{tab:abstract_retrieval} compares the performance of \emph{abstract retrieval} modules using using TF-IDF and BioSentVec \cite{chen2019biosentvec}. As Table \ref{tab:abstract_retrieval} indicates, the overall difference between these two methods is small. \citet{Wadden2020FactOF} chose $k_{retrieval}=3$ to maximize the $F_1$ score of the abstract retrieval module, while we choose a larger $k_{retrieval}$ to pursue a larger recall score, in order to retrieve more positive abstracts for the down-stream models.

\subsection{Baseline Models}

\paragraph{\textsc{VeriSci}.}
Along with the \textsc{SciFact} task and dataset, \citet{Wadden2020FactOF} proposed \textsc{VeriSci}, a sentence-level, pipeline-based solution. After retrieving the top similar abstracts for each claim with TF-IDF vectorization method, they applied a sentence-level ``BERT to BERT'' model \citet{deyoung2019eraser} to extract rationales, sentence by sentence, with a BERT model, and they predict the stance with another BERT model using the concatenation of the extracted rationale sentences. \citet{Wadden2020FactOF} used Roberta-large \cite{liu2019roberta} as their BERT model and pre-trained their \emph{stance prediction} module on the FEVER dataset \cite{thorne2018fever}.


\paragraph{\textsc{VerT5erini}.} Very recently, \citet{pradeep2020scientific} proposed a strong model \textsc{VerT5erini}, based on T5 \cite{raffel2019exploring}. They applied T5 for all three steps of the \textsc{SciFact} task in a sentence-level, pipeline fashion. Because of the known significant performance gap between Roberta-large \cite{liu2019roberta} that we use and T5 \cite{raffel2019exploring, pradeep2020scientific}, we only use \textsc{VerT5erini} as a reference (marked with *).

\subsection{Model Performances and Ablation Studies} 
We experiment on the oracle task, which performs \emph{rationale selection} and \emph{stance prediction} given the oracle abstracts (Table \ref{tab:oracle_dev}), and the open task, which performs the full task of \emph{abstract retrieval}, \emph{rationale selection}, and \emph{stance prediction} (Table \ref{tab:open_dev}). We tune our models based on the sentence-level, final development set performance (Selection+Label). The test labels are not released by \citet{Wadden2020FactOF}. Unless explicitly stated, all models are pre-trained on FEVER \cite{thorne2018fever}.

\paragraph{Paragraph-level Model vs. Sentence-level Model.}
We compare our paragraph-level pipeline model against \textsc{VeriSci} \cite{Wadden2020FactOF}, which is a sentence-level solution on the oracle task. As Table \ref{tab:oracle_dev} shows, our paragraph-level pipeline model (Paragraph-Pipeline) outperforms \textsc{VeriSci}, particularly on \emph{rationale selection}. This suggests the benefit of computing the contextualized sentence representations using the \textit{compact paragraph encoding} over individual sentence representations.

\paragraph{Joint Model vs. Pipeline Model.}
Although our joint model does not show benefits over the pipeline model on the oracle task (Table \ref{tab:oracle_dev}), the benefit emerges on the open task. Along with negative sampling, which greatly increases the tolerance of models to false positive abstracts, the Paragraph-Joint model shows its benefit over the Paragraph-Pipeline model. The small difference between the  Paragraph-Joint model and the same model except with TF-IDF \emph{abstract retrieval} (Paragraph-Joint TF-IDF) shows that the performance improvement is mainly attributed to the joint training, instead of replacing TF-IDF similarity with BioSentVec embedding similarity in \emph{abstract retrieval}.

\paragraph{Pre-training vs. Domain Adaptation.}
We also compare two methods of transfer learning from FEVER \cite{thorne2018fever} to \textsc{SciFact} \cite{Wadden2020FactOF}. 
Table \ref{tab:open_dev} shows that the effect of pre-training (Paragraph-Joint) or domain adaptation \cite{peng2016multi} (Paragraph-Joint DA) is similar. Both of them are effective as transfer learning, as they significantly outperform the same model that is only trained on \textsc{SciFact} (Paragraph-Joint \textsc{SciFact}-only).

\paragraph{KGAT  vs. Simple Attention as \emph{Stance Prediction} Module.}
We expected a significant performance improvement by applying the strong stance prediction model KGAT \cite{liu2020fine}, but the actual improvement is limited. This is likely due to the strong regularization of KGAT that under-fits the training data.

\paragraph{Test-set Performance on the \textsc{SciFact} Leaderboard}
By the time this paper is updated, our Paragraph-Joint model trained on the combination of \textsc{SciFact} training set
and development set achieved the first place on the \textsc{SciFact} leaderboard \footnote{\url{https://scifact.apps.allenai.org/leaderboard}, as of January 23, 2021.}. We obtain test sentence-level $F_1$ score (Selection+Label) of $60.9\%$ and test abstract-level $F_1$ score (Label+Rationale) of $67.2\%$.

\section{Related Work}
Fact-verification has been widely studied. There are many datasets available on various domains \cite{vlachos2014fact, ferreira2016emergent, popat2017truth, wang2017liar, derczynski2017semeval, popat2017truth, atanasova2018lluis, baly2018integrating,chen2019seeing,hanselowski2019richly}, among which the most influential one is FEVER shared task \cite{thorne2018fever}, which aims to develop systems to check the veracity of human-generated claims by extracting evidences from Wikipedia. Most existing systems \cite{nie2019combining} leverages a three-step pipeline approach by building modules for each of the step: document retrieval, rationale selection and fact verification. Many of them focus on the claim verification step \cite{zhou2019gear, liu2020fine}, such as KGAT \cite{liu2020fine}, one of the top models on FEVER leader board. On the other hand, there are some attempts on jointly optimizing rationale selection and stance prediction. TwoWingOS \cite{yin2018twowingos} leverages attentive CNN \cite{yin2018attentive} to inter-wire two modules, while \citet{hidey2020deseption} used a single pointer network \cite{vinyals2015pointer} for both sub-tasks. We propose another variation that directly links two modules by a dynamic attention mechanism.

Because \textsc{SciFact} \cite{Wadden2020FactOF} is a scientific version of FEVER \cite{thorne2018fever}, systems designed for FEVER can be applied to \textsc{SciFact} in principle. However, as a fact-verification task in scientific domain, \textsc{SciFact} task has inherited the common issue of lacking sufficient data, which can be mitigated with transfer learning by leveraging language models and introducing external dataset. The baseline model by \citet{Wadden2020FactOF} leverages Roberta-large \cite{liu2019roberta} fine-tuned on FEVER dataset \cite{thorne2018fever}, while \textsc{VerT5erini} \cite{pradeep2020scientific} leverages T5 \cite{raffel2019exploring} and fine-tuned on MS MARCO dataset \cite{bajaj2016ms}. In this work, in addition to fine-tuning Roberta-large on FEVER, we also explore domain adaptation \cite{peng2016multi} to mitigate the low resource issue.

\section{Conclusion}
In this work, we propose a novel paragraph-level multi-task learning model for \textsc{SciFact} task. Experiments show that (1) The compact paragraph encoding method is beneficial over separately computing sentence embeddings. (2) With negative sampling, the joint training of \emph{rationale selection} and \emph{stance prediction} is beneficial over the pipeline solution.

\section*{Acknowledgement}
We thank the anonymous reviewers for their useful comments, and Dr. Jessica Ouyang for her feedback. This work is supported by a National Institutes of Health (NIH) R01 grant (LM012592). The views and conclusions of this paper are those of the authors and do not reflect the official policy or position of NIH.

\clearpage
{\fontsize{9.0pt}{10.0pt} \selectfont
\bibliography{aaai2021}
}

\clearpage

\end{document}